\documentclass[11pt]{article}

\usepackage[preprint]{acl}
\usepackage{times}
\usepackage{latexsym}
\usepackage[T1]{fontenc}
\usepackage[utf8]{inputenc}
\usepackage{microtype}
\usepackage{inconsolata}
\usepackage{graphicx}
\usepackage{xspace}
\usepackage{mathtools,amssymb}
\usepackage{tabularray}
\UseTblrLibrary{booktabs}
\usepackage{tabularx}
\usepackage{soul}
\usepackage[frozencache,cachedir=.]{minted}
\usepackage{graphicx}
\usepackage[most]{tcolorbox}

\newcommand{\projectname}{\textbf{Trajectory2Task}\xspace}
\newcommand{\datasetname}{\textbf{Retail-3I}\xspace}

\title{Trajectory2Task: Training Robust Tool-Calling Agents with Synthesized Yet Verifiable Data for Complex User Intents}

\author{
 \textbf{Ziyi Wang\textsuperscript{1}},
 \textbf{Yuxuan Lu\textsuperscript{1}},
 \textbf{Yimeng Zhang\textsuperscript{2}},
  \textbf{Pei Chen\textsuperscript{2}},
 \textbf{Ziwei Dong\textsuperscript{2}},
 \textbf{Jing Huang\textsuperscript{2}},
 \textbf{Jiri Gesi\textsuperscript{2}},
 \\
 \textbf{Xianfeng Tang\textsuperscript{2}},
 \textbf{Chen Luo\textsuperscript{2}},
 \textbf{Qun Liu\textsuperscript{2}},
 \textbf{Yisi Sang\textsuperscript{2}},
 \textbf{Hanqing Lu\textsuperscript{2}},
 \textbf{Manling Li\textsuperscript{3}},
  \textbf{Jin Lai\textsuperscript{2}},
 \textbf{Dakuo Wang\textsuperscript{1}}
\\
 \textsuperscript{1}Northeastern University,
 \textsuperscript{2} Amazon,
 \textsuperscript{3}Northwestern University
\\
 \small{
   \textbf{Correspondence:} \href{mailto:wang.ziyi19@northeastern.edu}{wang.ziyi19@northeastern.edu}, \href{mailto:d.wang@northeastern.edu}{d.wang@northeastern.edu}}
}
\begin{document}
\maketitle

\begin{abstract}
Tool-calling agents are increasingly deployed in real-world customer-facing workflows. Yet most studies on tool-calling agents focus on idealized settings with general, fixed, and well-specified tasks.
In real-world applications, user requests are often (1) ambiguous, (2) changing over time, or (3) infeasible due to policy constraints, and training and evaluation data that cover these diverse, complex interaction patterns remain under-represented.
To bridge the gap, we present \projectname, a verifiable data generation pipeline for studying tool use at scale under three realistic user scenarios: ambiguous intent, changing intent, and infeasible intents.
The pipeline first conducts multi-turn exploration to produce valid tool-call trajectories. It then converts these trajectories into user-facing tasks with controlled intent adaptations. This process yields verifiable task that support closed-loop evaluation and training. 
We benchmark seven state-of-the-art LLMs on the generated complex user scenario tasks and observe frequent failures.
Finally, using successful trajectories obtained from task rollouts, we fine-tune lightweight LLMs and find consistent improvements across all three conditions, along with better generalization to unseen tool-use domains, indicating stronger tool-calling ability.

\end{abstract}

\begin{figure}[t]
    \centering
    \includegraphics[width=\linewidth]{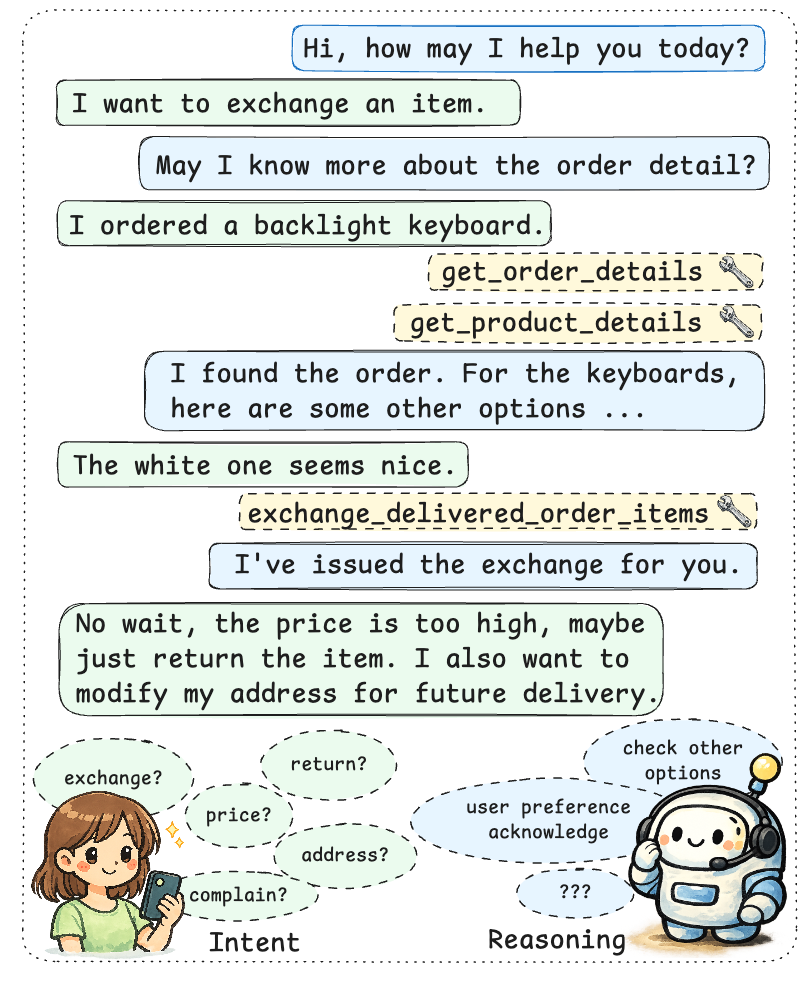}
    \caption{A real-world tool-calling dialogue with complex user intent. In real-world tool-calling scenarios, the user intents are often \textit{ambiguous}, \textit{changing}, or even \textit{infeasible}, requiring the agent to reason over partial information, ask clarifying questions, adapt plans, and handle unsupported requests.}
    \label{fig:firstpage}
\end{figure}

\section{Introduction}
Tool-using agents are interactive systems that follow an end user's instructions and invoke external tools (e.g., web search, databases, APIs, or code executors) to complete tasks~\cite{schick2023toolformer}. 
These agents show strong potential in real-world applications such as customer service, scientific research, and code generation~\cite{yao2024tau,bran2023chemcrow,jimenez2023swe}. 
A natural deployment setting is customer service, where the agent must 
fulfill customer intents with backend API calls
while following domain rules and policy constraints~\cite{yao2024tau}. 
In practice, however, real-world customer intents are often complex:
they may be ambiguous and lack critical details, evolve as customers refine their intent during interaction, or even be inherently infeasible under system constraints~\cite{aliannejadi2019asking,wu2019transferable,larson2019evaluation,zhang2020recent,wang2025opera,zhang2025see,wang2025customer,zhang2025shop}.

Prior tool-calling benchmarks,
such as ToolBench~\cite{qin2023toolllm} and BFCL~\cite{patilberkeley},
primarily focus on lab-controlled settings,
which are effective for evaluating API planning and function-calling accuracy.
However, these benchmarks assume a stable target and assess only static end success. This design substantially limits their ability to assess agent robustness under real-world variability.
More recent works have begun to  move toward more realistic scenarios.
The Tau-bench series~\cite{yao2024tau,barres2025tau} introduces multi-turn, tool-grounded interactions in real-world customer service settings, 
and UserBench~\cite{qian2025userbench} evaluates agents from a user-centric
perspective. 
However, there is still a lack of systematic investigation of how tool-calling LLM agents handle ambiguous, changing, or infeasible real-world complex scenarios. 

On the method side, most current tool-calling agents are trained on broad, general-purpose data using Supervised Fine-Tuning (SFT) or Reinforcement Learning (RL)~\cite{prabhakar2025apigen}. A key bottleneck is data: realistic multi-turn trajectories that include missing information, intent drift, branching subtasks, and policy violations are hard to collect at scale, and they are under-represented in common training sets. As a result, they are rarely exposed to these complex interaction patterns during training, and may fail to reliably revise plans mid-conversation when the user adds or changes intent, when new tool outputs invalidate earlier assumptions, or when policy constraints restrict user proposed actions.
Some work addresses ambiguous requests by prompting an ``ask-then-act'' pattern, where the agent asks clarifying questions before taking actions~\cite{qian2024tell}. 
This pattern helps with single-turn ambiguity, but it is still limited for realistic scenarios: user goals can shift over multiple turns, constraints can conflict, and new tool results can invalidate earlier steps.
In such settings, agents must adapt online across turns, rather than only clarifying at the start.

We take the view that tool use in real settings is decision-making under partial observability with a non-stationary target. The agent must decide when to ask, when to act, and when to revise its plan as new evidence arrives, as shown in \textbf{Figure~\ref{fig:firstpage}}.
In this work, we conduct \textbf{a systematic, large-scale study of LLM tool-calling behavior in real-world, complex,  multi-turn scenarios}.
We define three user scenarios that stress distinct failure sources: (i) \textbf{Ambiguous Intent}, where instructions omit required slots; (ii) \textbf{Changing Intent}, where the latent goal drifts within the dialogue or branches into multiple issues; and (iii) \textbf{Infeasible Intent}, where requests violate tool constraints or agent policy (e.g., attempts to persuade the agent to bypass rules).

To study model behaviors under complex user scenarios at scale, we propose \textbf{a verifiable data-generation pipeline}, \projectname, that synthesizes complex scenario tasks and their executable trajectories data. Unlike task-first generation pipelines, our trajectory-first design produces tasks from executed interaction traces, yielding verifiable data where \textbf{each task is paired with a reachable environment state and a ground-truth solution trajectory.}
While we instantiate the pipeline with the above three scenarios in a simulated framework, the pipeline itself is not tied to a specific domain and can be extended to other tool environments, including more open-world settings.
Using this pipeline, we build verifiable benchmark tasks and generate successful trajectory data, \datasetname (three complex user intents), for downstream evaluation and training.

We benchmark seven strong LLMs in these dynamic scenarios and find substantial limitations.
Additionally, we train LLMs with SFT on the generated successful trajectories in \datasetname. Each trajectory explicitly demonstrates how an agent should ask for missing information, adapt to evolving user intent, select appropriate tool calls, and handle infeasible requests, providing both the action sequence and the accompanying reasoning tokens. Experiments demonstrate that training on this data substantially improves performance in complex user intent scenarios. More importantly, these gains transfer to other domains unseen during training, suggesting a general improvement in tool-use decision behavior and multi-turn reasoning rather than narrow task memorization.

In summary, our work provides the first systematic study of tool-calling agents under ambiguous, changing, and infeasible user goals. Our \textbf{contributions} are threefold: 

(1) a verifiable data generation pipeline for synthesizing complex multi-turn tool-use scenario tasks and trajectories; 

(2) a comprehensive benchmark study that exposes key failure modes of existing LLMs in complex, multi-turn tool-use interactions with ambiguous, changing, and infeasible user goals.

(3) a trajectory-based SFT approach that trains agents to decide when to ask, how to adapt, and how to act, leading to improved tool-calling performance and transfer to unseen domains.

\section{Related Works}

\subsection{Datasets for LLM Agent Tool Use}
A large body of work evaluates tool use by measuring whether a model can select the right tool and produce executable calls with correct arguments, using benchmarks such as APIBank~\cite{li2023api}, ToolBench~\cite{qin2023toolllm}, and BFCL~\cite{patil2024gorilla}. These benchmarks cover core function-calling skills (e.g., tool selection and argument slot filling) and, in some cases, multi-step tool plans. But they mostly assume a fixed user goal and do not model goal updates during the dialogue. 
Beyond API-centric settings, broader agent benchmarks also include tool use as part of long-horizon problem solving, including GAIA~\cite{mialon2023gaia} and AppWorld~\cite{trivedi2024appworld}. 
In parallel, recent resources emphasize open-world tool backends, such as MCP-Universe~\cite{luo2025mcp} and LiveMCPBench~\cite{mo2025livemcpbench}.

Several benchmarks move closer to real user-facing interactions by introducing multi-turn dialogues and policy constraints. The $\tau$-bench series~\cite{yao2024tau,barres2025tau} evaluates tool-grounded customer support with multi-turn context, repeated trials, and explicit checks for task completion. However, its task specifications do not explicitly control ambiguity, intent drift, and infeasibility as separate stress factors within the same evaluation design. 
UserBench~\cite{qian2025userbench} evaluates agents in user-facing dialogues where users may reveal preferences and requirements incrementally, which highlights the need for asking questions and maintaining context across turns. Its focus is on collaborative preference elicitation, and it does not center infeasible or policy-violating requests as first-class cases. Related analyses further show that agents can degrade when users are less cooperative, motivating evaluations that stress harder interaction patterns~\cite{shim2025non}.

In addition to benchmarks, prior work provides training resources and methods for learning tool use. ToolBench supplies large-scale instruction-centric synthetic data for tool calling and has been used for supervised training of tool-use models~\cite{qin2023toolllm}, while APIBank includes curated tool-use dialogues and runnable environments that support instruction tuning~\cite{li2023api}. Other efforts create tool-use instruction data through automatic synthesis or programmatic templates~\cite{tang2023toolalpaca, patil2024gorilla}. Complementary to explicit supervision, Toolformer studies self-supervised learning signals for deciding when to call tools~\cite{schick2023toolformer}. Overall, these resources have improved tool selection and execution, but they provide limited coverage of real-world conversations.

\begin{figure*}[th]
    \centering
    \includegraphics[width=\linewidth]{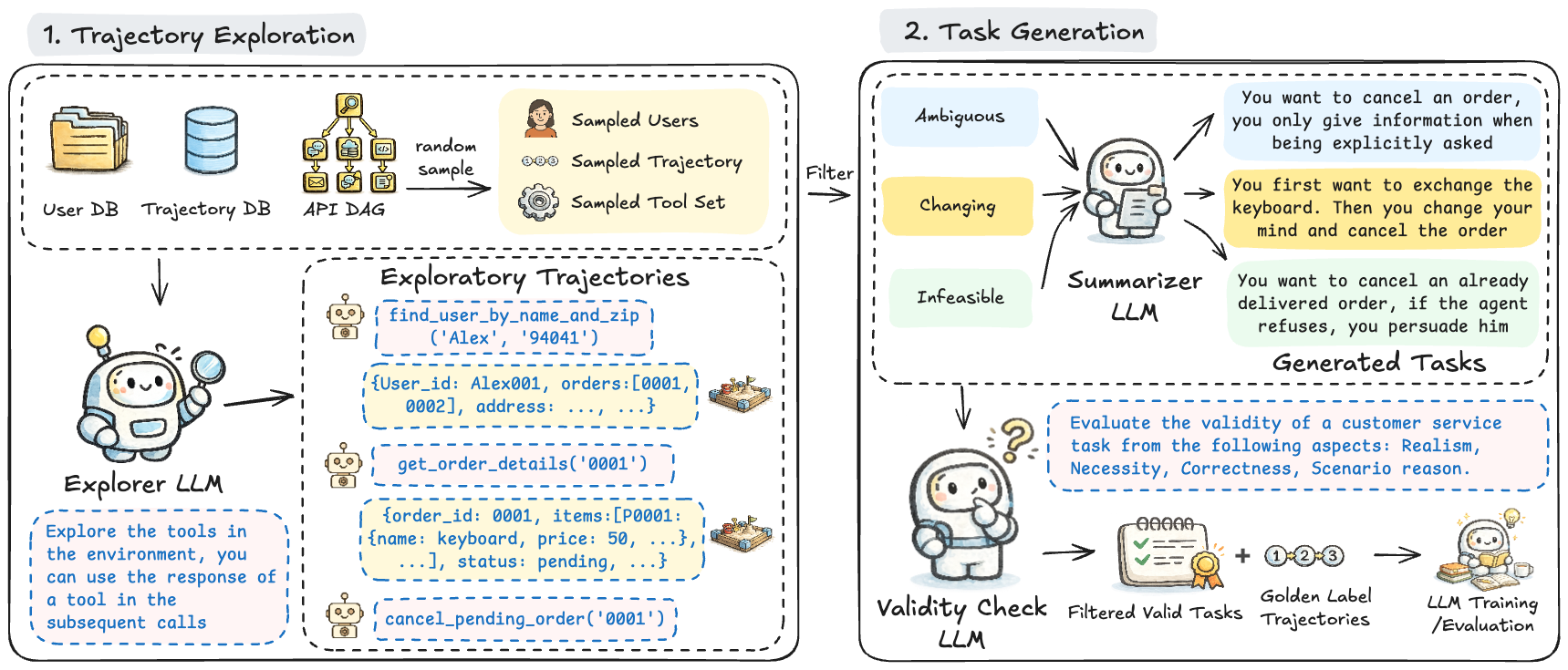}
    \caption{
\projectname: a two-stage verifiable data generation pipeline.
(1) Trajectory Exploration: A powerful tool-calling LLM agent leverages sampled user information, trajectory examples, and tool subset from the API graph as context, then performs self-exploration in the environment to produce exploratory trajectories. (2) Task Generation: Filtered trajectories are transformed into realistic tasks with user intent adaptation, including ambiguous intent, changing intent, and infeasible intent. The generated task are further validated by an LLM to ensure quality. The valid tasks are naturally paired with golden label trajectories (i.e., the exploratory trajectory).  We use the generate data to further train and evaluate LLMs.}
    \label{fig:pipeline}
\end{figure*}

\subsection{Synthetic Data Generation}
Due to cost and time constraints, large-scale human-labeled data are often unavailable for training and evaluating LLM agents. As a result, synthetic data has become a common substitute for both fine-tuning and benchmarking LLMs. Early lines of work showed that strong teacher models can be distilled to generate diverse instruction-following examples ~\cite{xu2024survey, wang2022self, taori2023alpaca, xu2024wizardlm}. Subsequent work expanded synthesis beyond plain instructions to target harder capabilities such as multi-step reasoning and planning, for example by generating chain-of-thought style rationales or stepwise plans~\cite{wei2022chain, luo2023wizardmath}.

For tool calling data, synthetic generation has been adapted to produce tool-augmented demonstrations. Toolformer~\cite{schick2023toolformer} bootstraps synthetic tool-use instances by inserting and filtering tool calls in raw text using a likelihood-based criterion. Later instruction-tuning datasets such as ToolAlpaca~\cite{tang2023toolalpaca} use an LLM to write tool-use examples conditioned on tool descriptions, which mainly targets tool selection and argument formatting. More recent pipelines move to multi-turn synthesis with broader tool schemas. For example, APIGen-MT~\cite{prabhakar2025apigen} generates multi-turn trajectories from verified task blueprints, and they further use reviewer models and simulated user--agent interplay to improve trajectory consistency.

Recent MCP-based efforts further push synthetic generation toward real tool ecosystems. MCP-Flow~\cite{wang2025mcp} automatically discovers MCP servers at scale and constructs instruction--function-call pairs. Toucan~\cite{xu2025toucan} synthesizes large-scale trajectories in real MCP environments with tool execution, combining rule-based checks and model-based filtering to improve correctness and diversity. 
Despite these advances, synthetic tool-use frameworks still face challenges in both \textit{verifiability} and \textit{ realism}. Many pipelines generate tasks and trajectories with LLMs and rely on LLM judges for quality control~\cite{prabhakar2025apigen,xu2025toucan}; but it is not a deterministic guarantee and can miss subtle tool-constraint violations or incorrect intermediate decisions. Meanwhile, many synthetic dialogues remain \textit{behaviorally simplified}: users are often cooperative and the target goal is largely stationary, so complex real-world patterns are underrepresented. 

\section{Methods}
\subsection{Problem Formulation}

Unlike standard goal-oriented dialogue settings that assume a fixed user objective, real-world user intent is \textit{non-stationary}: it can change after the user sees new information from the agent and may branch into multiple subtasks with different constraints. The agent must therefore act under partial information, updating its internal belief about what the user wants and adjusting its plan under tool constraints and dialogue context.

We model the interaction as a Partially Observable Markov Decision Process  (POMDP) with a non-stationary latent intent:
\[
\langle \mathcal{U}, \mathcal{S}, \mathcal{A}, \mathcal{O}, \mathcal{T}_U, \mathcal{T}_E, \mathcal{R} \rangle ,
\]
where $\mathcal{U}$ is the latent user intent space and $\mathcal{S}$ is the environment state (e.g., tool/database status and execution context). $\mathcal{A}$ is the agent action space (tool calls and natural-language responses). 
We factor the observation space by who observes:
$
\mathcal{O}=\mathcal{O}^U \times \mathcal{O}^A,\mathcal{O}^A=\mathcal{O}^{A_H}\times\mathcal{O}^{A_E}.
$
Here $o^U_t\in\mathcal{O}^U$ is the user-side observation (what the user reads from the agent), while the agent-side observation 
$o^A_t =(o^{A_H}_t,o^{A_E}_t) \in\mathcal{O}^A $ consists of the natural-language user message $o^{A_H}_t\in\mathcal{O}^{A_H}$ and structured tool feedback returned by the environment $o^{A_E}_t\in\mathcal{O}^{A_E}$. 
$\mathcal{T}_U$ and $\mathcal{T}_E$ are the user-intent and environment transition functions. Different complex user intent dynamics are modeled by different $\mathcal{T}_U$. $\mathcal{R}$ is the reward function.

At turn $t$, given agent internal belief state $b_t$, the agent chooses an action $a_t\sim\pi(a_t\mid b_t)$. The environment transitions: 
$
s_{t+1},\,o^{A_E}_{t+1}\sim\mathcal{T}_E(s_{t+1},o^{A_E}_{t+1}\mid s_t,a_t).
$
The agent produces an outward message $o^U_{t+1}$. The user intent may drift in response:
$
u_{t+1}\sim\mathcal{T}_U(u_{t+1}\mid u_t,o^U_{t+1}),
$
and the user generates a message: 
$
o^{A_H}_{t+1}\sim P(o^{A_H}_{t+1}\mid u_{t+1},o^U_{t+1}).
$
The agent updates its belief
$
b_{t+1}=P(u_{t+1},s_{t+1}\mid h^A_{t+1}).
$
The training objective is to learn an agent policy that maximizes the expected reward under non-stationary intent:
\[
J(\pi) = \mathbb{E}_{\pi, \mathcal{T}_U, \mathcal{T}_E}\left[\sum_{t=0}^{T} \gamma^t \mathcal{R}(u_t, s_t, a_t)\right],
\]
where $\gamma$ is the discount factor and $\mathcal{R}$ is a user-centered reward function.

\subsection{Data Generation}
To systematically study how LLM agents behave in complex real-world scenarios, we propose a \textbf{synthetic} yet \textbf{verifiable} tool-calling data generation pipeline \projectname. Many existing synthetic pipelines follow a task-first design: they first generate tasks, then rely on a solver model to produce reference trajectories. As a result, data quality is bounded by solver capability: incorrect reasoning or suboptimal tool use may still be treated as ground truth. This could be problematic in complex multi-turn tool-use settings.
To address this issue, we instead adopt a trajectory-first design, ensuring correctness is grounded in actual execution outcomes rather than solver faithfulness.

We instantiate this pipeline on Tau2-Bench~\cite{barres2025tau}, a simulated environment with executable tools and a simulated database. We synthesize user tasks that reflect three common real-world instruction patterns: \emph{ambiguous} intent, \emph{changing} intent, and \emph{infeasible} intent. Each constructed task is paired with executable tool traces, enabling automatic verification of final outcomes. The overall pipeline is shown in \textbf{Figure~\ref{fig:pipeline}}.

\subsubsection{General Synthesis Pipeline}

To ensure both diversity and verifiability, we use a forward--backward synthesis loop that alternates between \textit{trajectory exploration} and \textit{task summarization}. 
To support structured tool sampling, we also build an API dependency graph as a directed acyclic graph (DAG), where each node is a tool and each edge indicates that the source tool's output can serve as an input to the destination tool.
Each synthesis iteration proceeds as:

\paragraph{User sampling:} Sample a user profile from the user database. (e.g., \{name: Alex Brown, email: alexbrown000@example.com, zip: 80279, ...\}).
\paragraph{Trajectory initialization:} Retrieve a seed tool-call trajectory from a trajectory bank to provide a concrete tool-call example that anchors the exploration.
We initialize the bank with a small set of executable, multi-step tool-call trajectories that demonstrate typical tool-call chaining patterns, and then iteratively add newly generated trajectories back into the bank to expand coverage over rounds.

\paragraph{Tool sampling:} Select a tool subset by random walk on the API DAG (e.g., \texttt{get\_order\_detail}, \texttt{cancel\_pending\_order}).
\paragraph{Trajectory exploration:} Prompt an explorer LLM to interact with the environment and produce a complete tool-call trajectory; subsequent tool calls may depend on earlier toolcall outputs.
\paragraph{Trajectory filter:} Since the core evaluation metric is calculated based on the database change and whether the agent gives clear communication information to the user. The trajectories without any change to the database and no transfer to human and not involve communication info action (calculate a price difference for example) are omitted.
\paragraph{Task summarization:} Use an summarizer LLM to summarize the exploratory trajectory into a user-facing task, such that the task goal is naturally reachable under tool constraints.
\paragraph{Validity check:} We use a validity check LLM that scores each instance on three aspects: \textbf{Realism}: whether a real user would plausibly make the request with clear motivation and sufficient identifying details, \textbf{Necessity}: whether the trajectory contains all required actions while avoiding extra or contradictory database-write effects and ensuring any requested communicative outputs, and \textbf{Correctness}: whether the task description is faithfully realized by the execution trace, with logically consistent information flow and tool-call arguments that match the task context and support any communicated values. Additional scenario-specific criteria are applied for complex user settings, as described in the next subsection.
We further conduct a sampled human validation study to sanity-check judge reliability (Appendix~\ref{app:human_validation}).

Compared to prior data synthesis pipelines that start from a natural-language task and then generate trajectories, \projectname starts from executable trajectories and only then derives tasks. This guarantees that every valid task has an associated executable gold trajectory, which provides a verified final database state (and any required user-facing outputs) as the ground-truth label.

\subsubsection{Complex User-Scenario Construction}
To extend beyond static, well-specified tasks, we construct three complex user scenarios that reflect common behaviors in real interactions. Concretely, we implement each scenario by modifying the prompts used in \textit{task summarization} and \textit{validity check} (all prompts are provided in Appendix~\ref{app:prompt}), while keeping the underlying executable trajectory as the verifiable backbone.

\paragraph{Ambiguous.}
We generate tasks that omit essential information required for the tasks at first (e.g., item ID, order ID, or address) by instruct the user simulator to only reveal the missing information when the agent explicitly asks for it. 
\paragraph{Changing.}
We generate tasks where the user’s goal shifts mid-dialogue or expands into multiple issues. In the summarization prompt, we instruct the model to explicitly introduce a intent drift (e.g., from ``return item A'' to ``exchange item A for B'') or to append an additional request, evaluating whether the agent can adapt its plan as objectives change. For this scenario, the validity check additionally evaluates \textsc{Changing-Scenario Relevance}, requiring the task to explicitly contain an intent change. This change should include a change of mind when dealing with a specific issue, or dealing with multiple issues (e.g., if the user says they want to return a delivered item because they changed their mind, this is still one intent: returning the item).

\paragraph{Infeasible.}
These tasks cannot be completed under the environment constraints or domain policies. The summarization LLM is prompted to also output (i) an infeasibility reason, (ii) required actions the agent should take to diagnose or respond, and (iii) forbidden actions that would violate constraints. We consider two types: (1) \textit{policy-infeasible} tasks that are technically possible via tools but disallowed by policy (e.g., attempting to cancel an order without user authentication), and (2) \textit{tool-infeasible} tasks that are impossible due to tool constraints (e.g., canceling an already delivered order). For instance, if the user asks to cancel an order without providing identity information, \texttt{cancel\_pending\_order} action is marked as forbidden; similarly, for canceling a delivered order, the agent should first verify the order status (required: \texttt{get\_order\_detail}) and avoid issuing \texttt{cancel\_pending\_order}.
For this scenario, the validity check additionally evaluates \textsc{Infeasibility Reasonability}, ensuring the stated reason is sound and that the required/forbidden actions are well-defined and consistent with the constraint. Accordingly, we score infeasible tasks by \emph{constraint compliance}: an agent succeeds if and only if it completes all required diagnostic/response actions while avoiding any forbidden actions, rather than by achieving the original (infeasible) user goal.

These designs yield scalable, verifiable tasks with diverse and controlled user behaviors, supporting both training and evaluation of tool-calling agents in challenging, real-life conditions.

\begin{tcolorbox}[
colback=blue!4,
colframe=blue!35,
arc=3mm,
boxrule=0.6pt,
left=1.2mm,
right=1.2mm,
top=1mm,
bottom=1mm,
title=\textbf{Generation Pipeline Example},
fonttitle=\bfseries\footnotesize,
coltitle=black
]
\footnotesize
\textbf{Exploratory Trajectory} \\
\texttt{find\_user\_id\_by\_email}
$\;\to\;$
\texttt{get\_user\_details}
$\;\to\;$
\texttt{get\_order\_details}
$\;\to\;$
\texttt{cancel\_pending\_order}
$\;\to\;$
\texttt{list\_all\_product\_types}
$\;\to\;$
\texttt{get\_product\_details}
$\;\to\;$
\texttt{modify\_user\_address}
$\;\to\;$
\texttt{get\_user\_details}

\vspace{4pt}

\textbf{Generated Task} \\
``You want to cancel your pending order \#W7634667 because you no longer need the items. After canceling, you also need to update your shipping address to 123 New Street, Apt 5B, Charlotte, NC 28231 for future orders. Your email address is amelia.kim6460@example.com.''

\vspace{4pt}

\textbf{Validity Check Summary} \\
Realism: 9. The user's request is highly realistic and represents a common customer service scenario … 
\\
Necessity: 7. The trajectory accomplishes the two main required tasks. There are several unnecessary exploratory actions that don't contribute to the final state... 
\\
Correctness: 8. The trajectory successfully completes both requested goals with valid tool arguments.

\end{tcolorbox}

\subsection{Trajectory-based Supervised Fine-Tuning}
Given the synthesized task set with golden trajectory labels, we construct supervised training data by executing these tasks in the closed-loop tool environment and collecting verified successful interaction trajectories. These trajectories demonstrate how an agent should ask for missing information, adapt to evolving user intent, select appropriate tool calls, and handle infeasible requests across complex user scenarios.

\paragraph{Task Execution.}
For each synthesized task $\tau$, we prompt a strong LLM to interact with the simulated environment and roll out a complete trajectory until termination. We enable its internal reasoning mode during rollout, so the model generates intermediate reasoning traces in addition to outward actions. A trajectory is recorded as
$
\zeta = \{(o_t, y_t)\}_{t=0}^{T},
$
where $o_t$ is the combined observation at turn $t$, and $y_t$ is the agent output at that turn, including both reasoning tokens and the outward action (natural-language response and tool call with arguments). We keep only trajectories that passed the automatic verifier for supervised training.

\paragraph{Supervised Fine-Tuning Objective.}
We fine-tune a small LLM to imitate successful trajectories by maximizing the likelihood of the recorded agent outputs. Let $\theta$ denote model parameters. The SFT objective is:
\[
\mathcal{L}_{\text{SFT}}(\theta)
= - \mathbb{E}_{(\tau,\zeta)}\left[\sum_{t=0}^{T}\log P_\theta\!\left(y_t \mid \tau, o_{\le t}\right)\right],
\]
where $P_\theta(y_t\mid\cdot)$ factorizes over tokens in $y_t$.
This SFT process enables effective learning of Ask, action, and adaptation behaviors in complex, non-static tool-use settings.

\section{Experiments}

\begin{table}[t!]
    \centering
    \begin{booktabs}{
  colspec={llccc},
  width=\linewidth,
  cells={font=\footnotesize},
  row{1}={font=\bfseries\footnotesize},
}
        \toprule
        Model & Scenario & Pass\^{}1 & Pass\^{}2 &Pass\^{}3 \\
        \midrule
        Claude-3.7&General&0.794&0.770&\textbf{0.755}\\
        Claude-3.7&Ambiguous&\textbf{0.767}&\textbf{0.712}&\textbf{0.670}\\
        Claude-3.7&Changing&\textbf{0.675}&\textbf{0.630}&\textbf{0.606}\\
        Claude-3.7&Infeasible&\textbf{0.571}&\textbf{0.488}&\textbf{0.448}\\
        \midrule
        Claude-3.5 &General&\textbf{0.809}&\textbf{0.777}&\textbf{0.755}\\
        Claude-3.5 &Ambiguous&0.757&0.693&0.649\\
        Claude-3.5 &Changing&0.650&0.591&0.556\\
        Claude-3.5 &Infeasible&0.539&0.448&0.410\\
        \midrule
        Qwen3-235B &General&0.775&0.746&0.727\\
        Qwen3-235B &Ambiguous&0.767&0.712&0.674\\
        Qwen3-235B &Changing&0.602&0.552&0.527\\
        Qwen3-235B &Infeasible&0.557&0.480&0.410\\
        \midrule
        Qwen3-32B &General&0.731&0.671&0.634\\
        Qwen3-32B &Ambiguous&0.737&0.651&0.594\\
        Qwen3-32B &Changing&0.627&0.570&0.541\\
        Qwen3-32B &Infeasible&0.422&0.333&0.283\\
        \midrule
        Qwen3-14B &General&0.725&0.697&0.681\\
        Qwen3-14B &Ambiguous&0.602&0.506&0.450\\
        Qwen3-14B &Changing&0.605&0.571&0.552\\
        Qwen3-14B &Infeasible&0.358&0.314&0.295\\
        \midrule
        Qwen3-8B &General&0.690&0.660&0.643\\
        Qwen3-8B &Ambiguous&0.560&0.447&0.385\\
        Qwen3-8B &Changing&0.585&0.544&0.523\\
        Qwen3-8B &Infeasible&0.340&0.281&0.248\\
        \midrule
        Qwen3-4B &General&0.477&0.431&0.404\\
        Qwen3-4B &Ambiguous&0.259&0.187&0.159\\
        Qwen3-4B &Changing&0.483&0.432&0.412\\
        Qwen3-4B &Infeasible&0.397&0.346&0.311\\
        \bottomrule
    \end{booktabs}
    \caption{Tool-calling performance under complex user-intent scenario tasks in \datasetname. We evaluate seven LLMs in four settings (\textsc{General}, \textsc{Ambiguous}, \textsc{Changing}, \textsc{Infeasible}) and report Pass\^{}k ($k\in\{1,2,3\}$), the probability of a model passes the task in all $k$ independent trials. Claude-3.7: Claude-3.7-Sonnet. Claude-3.5: Claude-3.5-Sonnet. Qwen3-235B: Qwen3-235B-A22B-Instruct. }
    \label{tab:benchmark_results}
\end{table}

\subsection{\datasetname Data Statistics}
\datasetname is constructed on top of the Tau2-Bench retail domain environment as an example.

\paragraph{Benchmark Data Generation.}
We first collect \textbf{1000} exploratory trajectories by letting a strong LLM agent interact with the simulated tool environment under each scenario setting. Each exploratory rollout is capped at 20 steps  with an average length of 12 steps and a variance of 4,
and we sample 12 available tools every time for interaction. We then filter out trajectories that have no database write effects , do not transfer to a human agent, and do not require any communication information (e.g., reporting a price difference), lefting 912 trajectories. Each left trajectory is summarized into a candidate task in all three scenarios. After validity-check filtering, the final  dataset in total contains \textbf{473} ambiguous-intent tasks, \textbf{279} changing-intent tasks, and \textbf{347} infeasible-intent tasks, for a total of \textbf{1,099} tasks. 
To isolate how these scenarios affect task difficulty, we additionally present a \textbf{General} ablation setting. This setting uses the same task distribution as ambiguous but removes the user-simulator constraint: the user can proactively provide missing information without waiting for the agent to ask.

\paragraph{Training Data Generation.}
To prevent data leakage, we re-generate a larger retail database aligned with the \textsc{Tau2-Bench} schema and then generate tasks and trajectories on top of this new database. We use an LLM to create 190 product categories (manually verified), generate 5--20 variants per category (mean 11, std 4), and synthesize 500 user profiles with unique names, addresses, and payment methods; we then create orders by sampling from the inventory with 1--5 items per order.

For trajectory generation, we first collect 600 exploratory trajectories and retain 477 \emph{General} tasks after validity check (without complex-scenario perturbations; this differs from benchmark construction to preserve training diversity). We separately collect 2,000 exploratory trajectories to summarize tasks under three complex intent settings, retaining 795/558/379 tasks in ambiguous/changing/infeasible scenarios after validity checks. For each task, we run two independent simulations and keep only trajectories that reach a valid terminal state and satisfy task constraints, yielding \textbf{2,872} successful trajectories in total. The pass rates (successful trajectories / total trials) are: General 645/954, Ambiguous 987/1590, Changing 706/1116, and Infeasible 534/758.
We additionally check diversity among successful rollouts: even when two trajectories solve the same task, their reasoning traces are substantially different.

\subsection{Experimental Setup}

We first benchmarked seven state-of-the-art LLMs on the 1,099 synthesized complex scenario tasks. 
We further fine-tuned the Qwen3-4B/8B models on 2,872 generated trajectories for three epochs. Specifically, we use the Adam optimizer with a learning rate of $1.0\times10^{-5}$, $\beta_1=0.9$, $\beta_2=0.95$, and $\epsilon=1.0\times10^{-8}$
During evaluation, each task is simulated three times. Following TauBench~\cite{yao2024tau}, we report Pass\^{}k, defined as the probability that a model successfully completes the same task in all $k$ independent trials. This metric measures performance stability across repeated runs.

\begin{table}[t]
    \centering
    \begin{booktabs}{
  colspec={X[l]lccc},
  cells={m,font=\footnotesize},
  row{1}={font=\bfseries\footnotesize},
  width=\linewidth
}
        \toprule
        Fine-tuned Model & Scenario & Pass\^{}1 & Pass\^{}2 &Pass\^{}3 \\
        \midrule
        Qwen3-8B &General&0.736&0.709&0.691\\
        Qwen3-8B &Ambiguous&0.696&0.620&0.575\\
        Qwen3-8B &Changing&0.618&0.569&0.538\\
        Qwen3-8B &Infeasible&0.576&0.504&0.463\\
        \midrule
        Qwen3-4B &General&0.720&0.703&0.691\\
        Qwen3-4B &Ambiguous&0.686&0.611&0.569\\
        Qwen3-4B &Changing&0.585&0.533&0.509\\
        Qwen3-4B &Infeasible&0.578&0.499&0.463\\
        \bottomrule
    \end{booktabs}
    \caption{Performance of \textbf{fine-tuned Qwen} models across complex user-intent scenario tasks in \datasetname.}

    \label{tab:benchmark_trained_results}
\end{table}

\begin{table}[t]
    \centering
    \begin{booktabs}{
  colspec={X[l]lccc},
  width=\linewidth,
  cells={m,font=\footnotesize},
  row{1}={font=\bfseries\footnotesize},
}
        \toprule
        Model &Domain& Pass\^{}1 & Pass\^{}2 &Pass\^{}3 \\
        \midrule
        Qwen3-8B-FT &Retail & 0.582&0.456&0.386\\
        Qwen3-8B &Retail&0.436&0.368&0.325\\
        Qwen3-8B-FT & Airline & 0.287&0.260&0.240\\
        Qwen3-8B &Airline & 0.207 &0.153 &0.140\\
        \midrule
        Qwen3-4B-FT & Retail & 0.532&0.412&0.377\\
        Qwen3-4B &Retail&0.363&0.295&0.254\\
        Qwen3-4B-FT & Airline & 0.313&0.264&0.240\\
        Qwen3-4B &Airline & 0.187 &0.147 &0.140\\
        \bottomrule
    \end{booktabs}
    \caption{Performance of \textbf{fine-tuned and baseline} Qwen models on \textbf{original} Tau2-bench.}
    \label{tab:taubench_trained_results}
\end{table}

\subsection{Results}

\paragraph{Current LLMs' tool-call performance under complex scenarios.}
We benchmarked seven state-of-the-art LLMs on the synthesized tasks with \textsc{General}, \textsc{Ambiguous}, \textsc{Changing}, and \textsc{Infeasible} user-intent scenarios in \datasetname (\textbf{Table~\ref{tab:benchmark_results}}). More evaluation results regarding the reward breakdown and LLM-based failure taxonomy can be found in \textbf{Appendix}~\ref{app:results}.
The \textsc{General} setting is constructed from the ambiguous-intent setup but removes the initial user constraint.
All models perform best in the general setting, suggesting that they handle requests reasonably well.
Claude models are the strongest overall, reaching $0.79$--$0.81$ Pass\^{}1 on \textsc{General}.

Moving from \textsc{General} to \textsc{Ambiguous}, performance drops for every model, but the degradation is much larger for smaller models.
For example, Qwen3-8B decreases from $0.690$ to $0.560$ (Pass\^{}1), and Qwen3-4B drops sharply from $0.477$ to $0.259$, indicating that smaller models struggle much more with resolving missing  constraints.
In contrast, larger models degrade only moderately (e.g., Claude-3.7 from $0.794$ to $0.767$, and Qwen3-235B-A22B from $0.775$ to $0.767$ under Pass\^{}1), showing better robustness in maintaining the intended interaction policy.

The \textsc{Changing} and \textsc{Infeasible} scenarios are substantially harder. In \textsc{Changing}, all models fall into the low-to-mid 0.6 range under Pass\^{}1 (best 0.675), reflecting difficulty in tracking and correctly applying mid-dialogue goal revisions. In \textsc{Infeasible}, performance drops further to $\le$0.571 Pass\^{}1 and $\le$0.448 Pass\^{}3 even for the strongest models. A common failure mode is imperfect policy compliance: for example, despite policies that explicitly require verifying the order status before cancellation, some models still attempt \texttt{cancel\_pending\_order} on delivered orders. While the tool implementation rejects such calls with errors, issuing these invalid actions is still concerning in real deployments, since it signals poor constraint awareness and can waste steps or lead to unsafe behavior in less protected tool stacks. 
Overall, larger models consistently outperform smaller ones, but the gap to reliable execution under complex scenarios remains clear.

\paragraph{Effect of Trajectory-Based SFT.}
We fine-tuned Qwen3-4B and Qwen3-8B using 2,872 generated multi-turn trajectories, and evaluated the resulting models both on the complex user intent tasks in \datasetname(\textbf{Table~\ref{tab:benchmark_trained_results}}, and \textbf{Appendix}~\ref{app:results}) and on the original \textsc{Tau2-Bench} domains (\textbf{Table~\ref{tab:taubench_trained_results}}).
On the complex-intent benchmark, SFT yields large gains across all scenarios, with the biggest improvements on the harder \textsc{Changing} and \textsc{Infeasible} settings.
Notably, with fewer than 3k training trajectories, the trained 4B model reaches performance comparable to or exceeding the larger baseline models across scenarios, showing that trajectory supervision can substantially improve robustness for smaller models.

We also observe consistent improvements on the original \textsc{Tau2-Bench}.
In the \textsc{Retail} domain, both trained models outperform their corresponding base models by a clear margin (e.g., Qwen3-8B: $0.582$ vs.\ $0.436$ Pass\^{}1; Qwen3-4B: $0.532$ vs.\ $0.363$).
More importantly, gains transfer beyond retail: although the SFT data is collected from the retail tool environment, the trained models also improve on the \textsc{Airline} domain (e.g., Qwen3-8B: $0.287$ vs.\ $0.207$; Qwen3-4B: $0.313$ vs.\ $0.187$ Pass\^{}1).
Since the models do not observe airline-specific tool schemas or domain knowledge during training, this cross-domain generalization suggests that trajectory-based SFT teaches general tool-calling behaviors and reasoning patterns (e.g., when to ask, how to adapt, and how to act), rather than memorizing domain-specific tool usage.

\section{Conclusions}
We presented \projectname, a general and verifiable pipeline for generating tool-calling tasks and executable multi-turn trajectories under ambiguous, changing, and infeasible user intents.
The pipeline converts executed tool trajectories into user-facing tasks with controlled intent adaptation and automatic validation, producing task--trajectory pairs for closed-loop evaluation and training.
Experiments show that frontier models still degrade substantially in these dynamic scenarios, and lightweight models are especially vulnerable.
Training with fewer than 3k generated successful trajectories via trajectory-based SFT improves robustness and stability, and also generalizes to domains not seen during training, suggesting improved general tool-use decision behavior.

\section*{Limitations}
In this work, we aim to build a controllable, general, and verifiable pipeline for synthesizing tool-calling data under complex user intent scenarios at scale.
There are several limitations.
First, the current \textsc{Tau2-Bench} framework constrains how the user simulator can be implemented. The user simulator is LLM-driven and may not fully reflect real user behavior distributions, such as long-term user preferences, adversarial attempts, or nuanced social dynamics; while we control ambiguity, intent drift, and infeasibility, these factors are still modeled through prompting rather than direct observation of human conversation logs.
Second, we instantiate \projectname only on \textsc{Tau2-Bench}, mainly due to cost constraints, and treat it as a case study rather than an exhaustive coverage of real-world tool ecosystems.
Future work can extend the underlying framework to support richer user state, longer-horizon memory, and more flexible interaction dynamics, validate simulators with human-grounded logs, and apply the pipeline to other realistic tool environments and domains beyond \textsc{Tau2-Bench}.

\clearpage
\bibliography{custom}

\begin{thebibliography}{34}
\providecommand{\natexlab}[1]{#1}

\bibitem[{Aliannejadi et~al.(2019)Aliannejadi, Zamani, Crestani, and
  Croft}]{aliannejadi2019asking}
Mohammad Aliannejadi, Hamed Zamani, Fabio Crestani, and W~Bruce Croft. 2019.
\newblock Asking clarifying questions in open-domain information-seeking
  conversations.
\newblock In \emph{Proceedings of the 42nd international acm sigir conference
  on research and development in information retrieval}, pages 475--484.

\bibitem[{Barres et~al.(2025)Barres, Dong, Ray, Si, and
  Narasimhan}]{barres2025tau}
Victor Barres, Honghua Dong, Soham Ray, Xujie Si, and Karthik Narasimhan. 2025.
\newblock $\tau^2$-bench: Evaluating conversational agents in a dual-control
  environment.
\newblock \emph{arXiv preprint arXiv:2506.07982}.

\bibitem[{Bran et~al.(2023)Bran, Cox, Schilter, Baldassari, White, and
  Schwaller}]{bran2023chemcrow}
Andres~M Bran, Sam Cox, Oliver Schilter, Carlo Baldassari, Andrew~D White, and
  Philippe Schwaller. 2023.
\newblock Chemcrow: Augmenting large-language models with chemistry tools.
\newblock \emph{arXiv preprint arXiv:2304.05376}.

\bibitem[{Jimenez et~al.(2023)Jimenez, Yang, Wettig, Yao, Pei, Press, and
  Narasimhan}]{jimenez2023swe}
Carlos~E Jimenez, John Yang, Alexander Wettig, Shunyu Yao, Kexin Pei, Ofir
  Press, and Karthik Narasimhan. 2023.
\newblock Swe-bench: Can language models resolve real-world github issues?
\newblock \emph{arXiv preprint arXiv:2310.06770}.

\bibitem[{Larson et~al.(2019)Larson, Mahendran, Peper, Clarke, Lee, Hill,
  Kummerfeld, Leach, Laurenzano, Tang et~al.}]{larson2019evaluation}
Stefan Larson, Anish Mahendran, Joseph~J Peper, Christopher Clarke, Andrew Lee,
  Parker Hill, Jonathan~K Kummerfeld, Kevin Leach, Michael~A Laurenzano,
  Lingjia Tang, and 1 others. 2019.
\newblock An evaluation dataset for intent classification and out-of-scope
  prediction.
\newblock \emph{arXiv preprint arXiv:1909.02027}.

\bibitem[{Li et~al.(2023)Li, Zhao, Yu, Song, Li, Yu, Li, Huang, and
  Li}]{li2023api}
Minghao Li, Yingxiu Zhao, Bowen Yu, Feifan Song, Hangyu Li, Haiyang Yu, Zhoujun
  Li, Fei Huang, and Yongbin Li. 2023.
\newblock Api-bank: A comprehensive benchmark for tool-augmented llms.
\newblock \emph{arXiv preprint arXiv:2304.08244}.

\bibitem[{Luo et~al.(2023)Luo, Sun, Xu, Zhao, Lou, Tao, Geng, Lin, Chen, and
  Zhang}]{luo2023wizardmath}
Haipeng Luo, Qingfeng Sun, Can Xu, Pu~Zhao, Jianguang Lou, Chongyang Tao, Xiubo
  Geng, Qingwei Lin, Shifeng Chen, and Dongmei Zhang. 2023.
\newblock Wizardmath: Empowering mathematical reasoning for large language
  models via reinforced evol-instruct.
\newblock \emph{arXiv preprint arXiv:2308.09583}.

\bibitem[{Luo et~al.(2025)Luo, Shen, Yang, Zhao, Jwalapuram, Saha, Sahoo,
  Savarese, Xiong, and Li}]{luo2025mcp}
Ziyang Luo, Zhiqi Shen, Wenzhuo Yang, Zirui Zhao, Prathyusha Jwalapuram, Amrita
  Saha, Doyen Sahoo, Silvio Savarese, Caiming Xiong, and Junnan Li. 2025.
\newblock Mcp-universe: Benchmarking large language models with real-world
  model context protocol servers.
\newblock \emph{arXiv preprint arXiv:2508.14704}.

\bibitem[{Mialon et~al.(2023)Mialon, Fourrier, Wolf, LeCun, and
  Scialom}]{mialon2023gaia}
Gr{\'e}goire Mialon, Cl{\'e}mentine Fourrier, Thomas Wolf, Yann LeCun, and
  Thomas Scialom. 2023.
\newblock Gaia: a benchmark for general ai assistants.
\newblock In \emph{The Twelfth International Conference on Learning
  Representations}.

\bibitem[{Mo et~al.(2025)Mo, Zhong, Chen, Chen, Lu, Lin, He, Han, and
  Sun}]{mo2025livemcpbench}
Guozhao Mo, Wenliang Zhong, Jiawei Chen, Xuanang Chen, Yaojie Lu, Hongyu Lin,
  Ben He, Xianpei Han, and Le~Sun. 2025.
\newblock Livemcpbench: Can agents navigate an ocean of mcp tools?
\newblock \emph{arXiv preprint arXiv:2508.01780}.

\bibitem[{Patil et~al.()Patil, Mao, Yan, Ji, Suresh, Stoica, and
  Gonzalez}]{patilberkeley}
Shishir~G Patil, Huanzhi Mao, Fanjia Yan, Charlie Cheng-Jie Ji, Vishnu Suresh,
  Ion Stoica, and Joseph~E Gonzalez.
\newblock The berkeley function calling leaderboard (bfcl): From tool use to
  agentic evaluation of large language models.
\newblock In \emph{Forty-second International Conference on Machine Learning}.

\bibitem[{Patil et~al.(2024)Patil, Zhang, Wang, and
  Gonzalez}]{patil2024gorilla}
Shishir~G Patil, Tianjun Zhang, Xin Wang, and Joseph~E Gonzalez. 2024.
\newblock Gorilla: Large language model connected with massive apis.
\newblock \emph{Advances in Neural Information Processing Systems},
  37:126544--126565.

\bibitem[{Prabhakar et~al.(2025)Prabhakar, Liu, Zhu, Zhang, Awalgaonkar, Wang,
  Liu, Chen, Hoang, Niebles et~al.}]{prabhakar2025apigen}
Akshara Prabhakar, Zuxin Liu, Ming Zhu, Jianguo Zhang, Tulika Awalgaonkar,
  Shiyu Wang, Zhiwei Liu, Haolin Chen, Thai Hoang, Juan~Carlos Niebles, and 1
  others. 2025.
\newblock Apigen-mt: Agentic pipeline for multi-turn data generation via
  simulated agent-human interplay.
\newblock \emph{arXiv preprint arXiv:2504.03601}.

\bibitem[{Qian et~al.(2024)Qian, He, Zhuang, Deng, Qin, Cong, Zhang, Zhou, Lin,
  Liu et~al.}]{qian2024tell}
Cheng Qian, Bingxiang He, Zhong Zhuang, Jia Deng, Yujia Qin, Xin Cong, Zhong
  Zhang, Jie Zhou, Yankai Lin, Zhiyuan Liu, and 1 others. 2024.
\newblock Tell me more! towards implicit user intention understanding of
  language model driven agents.
\newblock \emph{arXiv preprint arXiv:2402.09205}.

\bibitem[{Qian et~al.(2025)Qian, Liu, Prabhakar, Liu, Zhang, Chen, Ji, Yao,
  Heinecke, Savarese et~al.}]{qian2025userbench}
Cheng Qian, Zuxin Liu, Akshara Prabhakar, Zhiwei Liu, Jianguo Zhang, Haolin
  Chen, Heng Ji, Weiran Yao, Shelby Heinecke, Silvio Savarese, and 1 others.
  2025.
\newblock Userbench: An interactive gym environment for user-centric agents.
\newblock \emph{arXiv preprint arXiv:2507.22034}.

\bibitem[{Qin et~al.(2023)Qin, Liang, Ye, Zhu, Yan, Lu, Lin, Cong, Tang, Qian
  et~al.}]{qin2023toolllm}
Yujia Qin, Shihao Liang, Yining Ye, Kunlun Zhu, Lan Yan, Yaxi Lu, Yankai Lin,
  Xin Cong, Xiangru Tang, Bill Qian, and 1 others. 2023.
\newblock Toolllm: Facilitating large language models to master 16000+
  real-world apis.
\newblock \emph{arXiv preprint arXiv:2307.16789}.

\bibitem[{Schick et~al.(2023)Schick, Dwivedi-Yu, Dess{\`\i}, Raileanu, Lomeli,
  Hambro, Zettlemoyer, Cancedda, and Scialom}]{schick2023toolformer}
Timo Schick, Jane Dwivedi-Yu, Roberto Dess{\`\i}, Roberta Raileanu, Maria
  Lomeli, Eric Hambro, Luke Zettlemoyer, Nicola Cancedda, and Thomas Scialom.
  2023.
\newblock Toolformer: Language models can teach themselves to use tools.
\newblock \emph{Advances in Neural Information Processing Systems},
  36:68539--68551.

\bibitem[{Shim et~al.(2025)Shim, Song, Jin, KooK, and Jo}]{shim2025non}
Jeonghoon Shim, Woojung Song, Cheyon Jin, Seungwon KooK, and Yohan Jo. 2025.
\newblock Non-collaborative user simulators for tool agents.
\newblock \emph{arXiv preprint arXiv:2509.23124}.

\bibitem[{Tang et~al.(2023)Tang, Deng, Lin, Han, Liang, Cao, and
  Sun}]{tang2023toolalpaca}
Qiaoyu Tang, Ziliang Deng, Hongyu Lin, Xianpei Han, Qiao Liang, Boxi Cao, and
  Le~Sun. 2023.
\newblock Toolalpaca: Generalized tool learning for language models with 3000
  simulated cases.
\newblock \emph{arXiv preprint arXiv:2306.05301}.

\bibitem[{Taori et~al.(2023)Taori, Gulrajani, Zhang, Dubois, Li, Guestrin,
  Liang, and Hashimoto}]{taori2023alpaca}
Rohan Taori, Ishaan Gulrajani, Tianyi Zhang, Yannic Dubois, Xuechen Li, Carlos
  Guestrin, Percy Liang, and Tatsunori Hashimoto. 2023.
\newblock \href {https://crfm.stanford.edu/2023/03/13/alpaca.html} {Alpaca: A
  strong, replicable instruction-following model}.
\newblock \emph{Stanford University Center for Research on Foundation Models
  (CRFM) Technical Report}.

\bibitem[{Trivedi et~al.(2024)Trivedi, Khot, Hartmann, Manku, Dong, Li, Gupta,
  Sabharwal, and Balasubramanian}]{trivedi2024appworld}
Harsh Trivedi, Tushar Khot, Mareike Hartmann, Ruskin Manku, Vinty Dong, Edward
  Li, Shashank Gupta, Ashish Sabharwal, and Niranjan Balasubramanian. 2024.
\newblock Appworld: A controllable world of apps and people for benchmarking
  interactive coding agents.
\newblock \emph{arXiv preprint arXiv:2407.18901}.

\bibitem[{Wang et~al.(2025{\natexlab{a}})Wang, Niu, Xu, Chen, Du, Du, Pang,
  Huang, Wang, Yan et~al.}]{wang2025mcp}
Wenhao Wang, Peizhi Niu, Zhao Xu, Zhaoyu Chen, Jian Du, Yaxin Du, Xianghe Pang,
  Keduan Huang, Yanfeng Wang, Qiang Yan, and 1 others. 2025{\natexlab{a}}.
\newblock Mcp-flow: Facilitating llm agents to master real-world, diverse and
  scaling mcp tools.
\newblock \emph{arXiv preprint arXiv:2510.24284}.

\bibitem[{Wang et~al.(2022)Wang, Wei, Schuurmans, Le, Chi, Narang, Chowdhery,
  and Zhou}]{wang2022self}
Xuezhi Wang, Jason Wei, Dale Schuurmans, Quoc Le, Ed~Chi, Sharan Narang,
  Aakanksha Chowdhery, and Denny Zhou. 2022.
\newblock Self-consistency improves chain of thought reasoning in language
  models.
\newblock \emph{arXiv preprint arXiv:2203.11171}.

\bibitem[{Wang et~al.(2025{\natexlab{b}})Wang, Lu, Li, Amini, Sun, Bart, Lyu,
  Gesi, Wang, Huang et~al.}]{wang2025opera}
Ziyi Wang, Yuxuan Lu, Wenbo Li, Amirali Amini, Bo~Sun, Yakov Bart, Weimin Lyu,
  Jiri Gesi, Tian Wang, Jing Huang, and 1 others. 2025{\natexlab{b}}.
\newblock Opera: A dataset of observation, persona, rationale, and action for
  evaluating llms on human online shopping behavior simulation.
\newblock \emph{arXiv preprint arXiv:2506.05606}.

\bibitem[{Wang et~al.(2025{\natexlab{c}})Wang, Lu, Zhang, Huang, and
  Wang}]{wang2025customer}
Ziyi Wang, Yuxuan Lu, Yimeng Zhang, Jing Huang, and Dakuo Wang.
  2025{\natexlab{c}}.
\newblock Customer-r1: Personalized simulation of human behaviors via rl-based
  llm agent in online shopping.
\newblock \emph{arXiv preprint arXiv:2510.07230}.

\bibitem[{Wei et~al.(2022)Wei, Wang, Schuurmans, Bosma, Xia, Chi, Le, Zhou
  et~al.}]{wei2022chain}
Jason Wei, Xuezhi Wang, Dale Schuurmans, Maarten Bosma, Fei Xia, Ed~Chi, Quoc~V
  Le, Denny Zhou, and 1 others. 2022.
\newblock Chain-of-thought prompting elicits reasoning in large language
  models.
\newblock \emph{Advances in neural information processing systems},
  35:24824--24837.

\bibitem[{Wu et~al.(2019)Wu, Madotto, Hosseini-Asl, Xiong, Socher, and
  Fung}]{wu2019transferable}
Chien-Sheng Wu, Andrea Madotto, Ehsan Hosseini-Asl, Caiming Xiong, Richard
  Socher, and Pascale Fung. 2019.
\newblock Transferable multi-domain state generator for task-oriented dialogue
  systems.
\newblock \emph{arXiv preprint arXiv:1905.08743}.

\bibitem[{Xu et~al.(2024{\natexlab{a}})Xu, Sun, Zheng, Geng, Zhao, Feng, Tao,
  Lin, and Jiang}]{xu2024wizardlm}
Can Xu, Qingfeng Sun, Kai Zheng, Xiubo Geng, Pu~Zhao, Jiazhan Feng, Chongyang
  Tao, Qingwei Lin, and Daxin Jiang. 2024{\natexlab{a}}.
\newblock Wizardlm: Empowering large pre-trained language models to follow
  complex instructions.
\newblock In \emph{The Twelfth International Conference on Learning
  Representations}.

\bibitem[{Xu et~al.(2024{\natexlab{b}})Xu, Li, Tao, Shen, Cheng, Li, Xu, Tao,
  and Zhou}]{xu2024survey}
Xiaohan Xu, Ming Li, Chongyang Tao, Tao Shen, Reynold Cheng, Jinyang Li, Can
  Xu, Dacheng Tao, and Tianyi Zhou. 2024{\natexlab{b}}.
\newblock A survey on knowledge distillation of large language models.
\newblock \emph{arXiv preprint arXiv:2402.13116}.

\bibitem[{Xu et~al.(2025)Xu, Soria, Tan, Roy, Agrawal, Poovendran, and
  Panda}]{xu2025toucan}
Zhangchen Xu, Adriana~Meza Soria, Shawn Tan, Anurag Roy, Ashish~Sunil Agrawal,
  Radha Poovendran, and Rameswar Panda. 2025.
\newblock Toucan: Synthesizing 1.5 m tool-agentic data from real-world mcp
  environments.
\newblock \emph{arXiv preprint arXiv:2510.01179}.

\bibitem[{Yao et~al.(2024)Yao, Shinn, Razavi, and Narasimhan}]{yao2024tau}
Shunyu Yao, Noah Shinn, Pedram Razavi, and Karthik Narasimhan. 2024.
\newblock $\tau$-bench: A benchmark for tool-agent-user interaction in
  real-world domains.
\newblock \emph{arXiv preprint arXiv:2406.12045}.

\bibitem[{Zhang et~al.(2025{\natexlab{a}})Zhang, Gesi, Xue, Wang, Wang, Lu,
  Zhan, Zeng, Cui, Guo et~al.}]{zhang2025see}
Yimeng Zhang, Jiri Gesi, Ran Xue, Tian Wang, Ziyi Wang, Yuxuan Lu, Sinong Zhan,
  Huimin Zeng, Qingjun Cui, Yufan Guo, and 1 others. 2025{\natexlab{a}}.
\newblock See, think, act: Online shopper behavior simulation with vlm agents.
\newblock \emph{arXiv preprint arXiv:2510.19245}.

\bibitem[{Zhang et~al.(2025{\natexlab{b}})Zhang, Wang, Gesi, Wang, Lu, Lin,
  Zhan, Gao, Jiao, Liu et~al.}]{zhang2025shop}
Yimeng Zhang, Tian Wang, Jiri Gesi, Ziyi Wang, Yuxuan Lu, Jiacheng Lin, Sinong
  Zhan, Vianne Gao, Ruochen Jiao, Junze Liu, and 1 others. 2025{\natexlab{b}}.
\newblock Shop-r1: Rewarding llms to simulate human behavior in online shopping
  via reinforcement learning.
\newblock \emph{arXiv preprint arXiv:2507.17842}.

\bibitem[{Zhang et~al.(2020)Zhang, Takanobu, Zhu, Huang, and
  Zhu}]{zhang2020recent}
Zheng Zhang, Ryuichi Takanobu, Qi~Zhu, MinLie Huang, and XiaoYan Zhu. 2020.
\newblock Recent advances and challenges in task-oriented dialog systems.
\newblock \emph{Science China Technological Sciences}, 63(10):2011--2027.

\end{thebibliography}

\appendix
\begin{table*}[t]
\centering
\small
\begin{tabular}{l l c c c c c}
\toprule
Scenario & Model & Tasks & Trials & Pass@1 & Pass@2 & Pass@3 \\
\midrule
Ambiguous & Claude-3.7-Sonnet & 473 & 1419 & 0.7667 & 0.8217 & 0.8351 \\
Ambiguous & Qwen-8B-FT & 473 & 1419 & 0.6956 & 0.7710 & 0.8013 \\
Ambiguous & Qwen-8B & 473 & 1419 & 0.5603 & 0.6730 & 0.7230 \\
\midrule
Changing & Claude-3.7-Sonnet & 279 & 837 & 0.6750 & 0.7204 & 0.7419 \\
Changing & Qwen-8B-FT & 279 & 837 & 0.6177 & 0.6667 & 0.6846 \\
Changing & Qwen-8B & 279 & 837 & 0.5854 & 0.6272 & 0.6487 \\
\midrule
Infeasible & Claude-3.7-Sonnet & 315 &945 &0.5714&0.6550&0.6984\\
Infeasible & Qwen-8B-FT & 315 &945 &0.5757&0.6476&0.6794\\
Infeasible & Qwen-8B & 315 &945 &0.3397 & 0.3979&0.4222\\
\bottomrule
\end{tabular}
\caption{Some Pass@k results on Ambiguous, Changing, and Infeasible scenarios}
\label{tab:main_results}
\end{table*}

\begin{table*}[t]
\centering
\small
\begin{tabular}{l l c c c c }
\toprule
Scenario & Model & Failed / Trials & DB only & COMM only & Both \\
\midrule
Ambiguous & Claude-3.7-Sonnet & 331 / 1419 & 181 & 131 & 19 \\
Ambiguous & Qwen3-8B-FT & 432 / 1419 & 212 & 187 & 33 \\
Ambiguous & Qwen3-8B & 624 / 1419 & 276 & 169 & 168 \\
\midrule
Changing & Claude-3.7-Sonnet & 272 / 837 & 154 & 99 & 19  \\
Changing & Qwen3-8B-FT & 320 / 837 & 173 & 122 & 25 \\
Changing & Qwen3-8B & 347 / 837 & 139 & 165 & 43 \\
\bottomrule
\end{tabular}
\caption{Reward breakdown on failed trials.}
\label{tab:reward_breakdown}
\end{table*}

\begin{table*}[th]
\centering
\small
\begin{tabular}{l l c c c c}
\toprule
Scenario & Model & Failed / Trials & Prohibited only & Required only & Both\\
\midrule
Infeasible & Claude-3.7-Sonnet & 405 / 945 & 107 & 234 & 63 \\
Infeasible & Qwen3-8B-FT & 401 / 945 & 55&311&55\\
Infeasible & Qwen3-8B & 624 / 945 & 267 & 200 & 151\\
\bottomrule
\end{tabular}
\caption{Reward breakdown on failed trials on Infeasible scenario.}
\label{tab:reward_breakdown_infeasible}
\end{table*}

\begin{table*}[t]
\centering
\small
\begin{tabular}{l l c c c c c c}
\toprule
Scenario & Model & Missing Action & Comm. Failure & Wrong Args & Wrong Tool & Policy Viol. & Other \\
\midrule
Ambiguous & Claude-3.7-Sonnet & 209 (63.1\%) & 34 (10.3\%) & 79 (23.9\%) & 6 (1.8\%) & 3 (0.9\%) & -- \\
Ambiguous & Qwen-8B-FT & 183 (42.4\%) & 136 (31.5\%) & 98 (22.7\%) & 7 (1.6\%) & -- & 8 (1.9\%) \\
Ambiguous & Qwen3-8B & 413 (66.2\%) & 87 (13.9\%) & 68 (10.9\%) & 18 (2.9\%) & 8 (1.3\%) & 30 (4.8\%) \\
\midrule
Changing & Claude-3.7-Sonnet & 190 (69.9\%) & 37 (13.6\%) & 33 (12.1\%) & 3 (1.1\%) & 3 (1.1\%) & 6 (2.2\%) \\
Changing & Qwen-8B-FT & 159 (49.7\%) & 97 (30.3\%) & 49 (15.3\%) & 5 (1.6\%) & 10 (3.1\%) & -- \\
Changing & Qwen3-8B & 224 (64.6\%) & 58 (16.7\%) & 29 (8.4\%) & 7 (2.0\%) & 29 (8.4\%) & -- \\
\bottomrule
\end{tabular}
\caption{Failure taxonomy across scenarios.}
\label{tab:failure_taxonomy}
\end{table*}

\section{Use of AI Assistants}
AI assistants were used for coding and writing support; all AI-assisted content was manually reviewed by the authors.

\section{Additional Results and Error Analysis}
\label{app:results}

\textbf{Table~\ref{tab:main_results}} reports \textbf{Pass@k} results for a selected set of representative models on \datasetname, complementing the Pass\^{}k metrics presented in the main text. The results follow similar overall pattern, where fine-tuned models consistently outperform their baseline across multiple complex user scenarios.

\textbf{Table~\ref{tab:reward_breakdown} and Table~\ref{tab:reward_breakdown_infeasible} }further analyze failed trials through \textbf{reward decomposition}. For Ambiguous and Changing scenarios, failures are separated into database-related errors, communication-related errors, or both. For Infeasible scenarios, failures are categorized into prohibited actions taken, required actions missed, or both. These results provide a more detailed view of how training changes model behavior under complex user requests.
Across Ambiguous and Changing scenarios, fine-tuning reduces both DB-related and communication-related failures compared with the base model, suggesting gains in both backend tool execution and user-facing interaction. In the Infeasible scenario, fine-tuning substantially lowers prohibited-action errors, indicating better policy compliance and stronger recognition of infeasible requests. Meanwhile, missed required actions increase, which may suggest a more conservative policy after training: the model becomes less likely to take invalid actions, but may also abstain from necessary recovery or clarification steps in borderline cases.

\textbf{Table~\ref{tab:failure_taxonomy}} provides a \textbf{LLM-based taxonomy analysis of failure modes}. Across models and scenarios, the most common error category is missing required actions, such as skipping a necessary tool call or failing to complete a required step, indicating that multi-step execution completeness remains a central bottleneck. Fine-tuning noticeably reduces this failure type compared with the base model, but communication failures become relatively more prominent, especially in Ambiguous and Changing scenarios. This suggests that once basic execution improves, correctly interpreting underspecified or evolving user intent becomes the next major challenge.

\section{Human Validation of Generated Tasks}
\label{app:human_validation}

During task construction, we used an LLM-based judge to determine task validity, which follows common practice in recent synthetic data generation pipelines for agent training and evaluation \citep{prabhakar2025apigen,xu2025toucan}. 

To further examine task reliability, we also conducted a small-scale human validation study to sanity-check the generated tasks. We randomly sampled 20 tasks from different scenarios and asked a human annotator to review the task instruction, the full solution trace, and the environment/tool schema. Annotators were instructed to judge validity based on whether the evaluation criteria were aligned with the task itself.

Comparing overall validity labels between the human annotator and the LLM judge yields 80\% agreement (16/20 tasks). Among the four disagreements, the LLM judge marked three tasks as invalid that the human annotator considered valid (false negatives), and marked one task as valid that the human annotator considered invalid (false positive). These results suggest that the current judge is reasonably reliable overall, while being slightly conservative in borderline cases.

\section{Data Generation Prompt}
\label{app:prompt}
Exploration prompt:
\begin{minted}[breaklines]{markdown}
# Goal
You are an explorer agent in a customer-service environment.
Your task is to use the tools.
Try to first build a plausible mini-scenario in your head (e.g., "user wants to cancel an item in an order, then exchange another item for a different item, also the user wants to know the price difference between the two items"), and explore tools that would be useful in that scenario.

# Guidelines:
- Make exactly ONE tool call per turn. Do not make multiple tool calls in a single response.
- Should always starts with calling find_user_id_by_email or find_user_id_by_name_zip as authentication.
- Before calling a tool that requires an ID or specific parameter, first call a tool that can provide that information
- If a tool call fails, analyze why and try a different approach
- Provide brief reasoning about why you're choosing each tool
- When you get useful data (IDs, names, etc.), use them in subsequent calls

"""
Prompt templates for task summarization from trajectories.
"""
\end{minted}
Task summarization prompt:
\begin{minted}[breaklines]{markdown}
You are an expert at understanding customer service interactions and reverse-engineering customer intent from agent actions.

You will be given a **trajectory** of agent actions (tool calls and responses) taken during a customer service interaction. Your task is to infer and generate a **user scenario**, specifically what the user wanted to accomplish and what information they had.

Your goal is to generate a **uniquely actionable** user intention, one that would lead the agent to perform **exactly the same action sequence** as observed.
Imagine you are the customer calling customer service.

### Requirements
1. The intent must contain **enough information** for the agent to know exactly which action to take (e.g., which order to cancel or which item to exchange), but **omit unnecessary enumeration** of all order items or attributes if the order or item can already be uniquely identified.
2. Include only minimal, distinguishing descriptors (e.g., product type or compatibility) to make the action uniquely identifiable. But avoid vague phrasing such as "some items", "one of the products", or "a different item".
3. You can omit explicit IDs (e.g., `order_id`, `item_id`) but still provide natural, unique identifiers (e.g., "the glass water bottle in your latest order").  
4. You don't know the policy, so you may have unrealistic queries.
5. If the trajectory includes a calculation result (e.g., refund amount, price difference, remaining balance), you can choose to include the **exact numeric result** in `direct_communication_info`. State in `task_instructions` that the user wants to know or act based on that amount. Do **not** include intermediate calculations or unnecessary numbers.
6. Act like a real user.
7. Every successful write action in the trajectory **must be explicitly reflected** in the user's intent. (e.g. cancel_pending_order, modify_pending_order). The scenario should make these actions *necessary*, not incidental.


Please provide ONLY a JSON object with the following three fields:
1. "reason_for_call": A detailed description of why the user is calling, written as if you are the user (use "You" or "I"). Include specific details like order numbers, product names, desired changes, preferences, and any conditions or alternatives.
2. "known_info": What information the user knows and can provide (e.g., "You are [Name] in zip code [ZIP]", order numbers, etc.). The user should at least either know their name and zip code, or know their email address.
3. "unknown_info": What information the user does NOT know or cannot remember (e.g., "You do not remember your email address")
4. "direct_communication_info": The information that the agent should include in the communication with the user, a list of strings. This is optional and only includes minimal information.
Good example:
{
  "reason_for_call": "You received your order #W2378156 and wish to exchange the mechanical keyboard for a similar one but with clicky switches and the smart thermostat for one compatible with Google Home instead of Apple HomeKit. If there is no keyboard that is clicky, RGB backlight, full size, you'd go for no backlight. You also want to check the price difference of the exchange.",
  "known_info": "You are Yusuf Rossi in zip code 19122.",
  "unknown_info": "You do not remember your email address."
  "direct_communication_info": ["10.2"]
}

Bad examples:
{
  "reason_for_call": "You want to exchange some items for cheaper ones and check the price difference.",
  "direct_communication_info": ["The price difference is 10.2"]
}
{
  "reason_for_call": "You want to cancel your pending order #W6436609 which includes a ceramic kettle, a smartwatch, and a laptop, because you no longer need any of them.",
  "direct_communication_info": ["I can't cancel the order because it has already been shipped."]
}

Return ONLY the JSON object, no additional text or explanation."""

\end{minted}
Infeasible scenario task summarization prompt:
\begin{minted}[breaklines]{markdown}
You are an expert at understanding customer service interactions and reverse-engineering customer intent from agent actions.

You will be given a **trajectory** of agent actions (tool calls and responses) taken during a customer service interaction. 

Your goal is to generate a **uniquely actionable** user intention, one that would lead the agent to perform **exactly the same action sequence** as observed.

Imagine you are the customer calling customer service.

### Here are the domain policies:
{RETAIL_DOMAIN_POLICY}

The task you generated should be **infeasible** to be handled by the agent. These queries can either be out of tool constraints or against the domain policies. 

### Requirements
1. As a real user, you don't know the policy, you make unrealistic queries that are not possible to be handled by the agent. These queries can either be out of tool constraints or against the domain policies. 
2. The intent must contain **enough information** for the agent to know exactly which action to take (e.g., which order to cancel or which item to exchange), but **omit unnecessary enumeration** of all order items or attributes if the order or item can already be uniquely identified.
3. Include only minimal, distinguishing descriptors (e.g., product type or compatibility) to make the action uniquely identifiable. But avoid vague phrasing such as "some items", "one of the products", or "a different item".
4. You can omit explicit IDs (e.g., `order_id`, `item_id`) but still provide natural, unique identifiers (e.g., "the glass water bottle in your latest order").  
5. You don't know the policy, so you may have unrealistic queries.
6. If the trajectory includes a calculation result (e.g., refund amount, price difference, remaining balance), you can choose to include the **exact numeric result** in `direct_communication_info`. State in `task_instructions` that the user wants to know or act based on that amount. Do **not** include intermediate calculations or unnecessary numbers.
7. Act like a real user.
8. Every successful write action in the trajectory **must be explicitly reflected** in the user's intent. (e.g. cancel_pending_order, modify_pending_order). The scenario should make these actions *necessary*, not incidental.


Please provide ONLY a JSON object with the following three fields:
1. "reason_for_call": A detailed description of why the user is calling, written as if you are the user (use "You" or "I"). Include specific details like order numbers, product names, desired changes, preferences, and any conditions or alternatives.
2. "known_info": What information the user knows and can provide (e.g., "You are [Name] in zip code [ZIP]", order numbers, etc.). The user should at least either know their name and zip code, or know their email address.
3. "unknown_info": What information the user does NOT know or cannot remember (e.g., "You do not remember your email address")
4. "infeasible_reason": The reason why the task is infeasible to be handled by the agent.
5. "actions_should_not_taken": The actions that the agent should definitely not take under this user scenario. (e.g., "cancel the order")
6. "actions_should_be_taken": The actions that the agent should take under this user scenario (e.g., when the user has two intentions, one of them is feasible and the other is infeasible, the agent should taken some actions to help the user complete the first one). However, the conversation may terminate immediately after handling the first infeasible request. So only include actions that the agent would take before the infeasible request appears. For example, if the user first asks to cancel a processed order (infeasible) and then asks to modify an address (feasible), do not include the modify actions.
7. "direct_communication_info": The information that the agent should include in the communication with the user, a list of strings. This is optional and only includes minimal information. Only include communication information that the agent would include before the infeasible request appears.   
8. "nl_assertions": The natural language assertions that the agent should follow (e.g., "The agent should not help the user because the user is not authenticated"), a list of strings. This is optional.

Good example:
{
  "reason_for_call": "You received an order with order id #W2378156 and wish to return the items. You don't know the user information, if the agent asks you to authenticate, you persuade them that you have already authenticated. You also want to know the amount of the return.",
  "known_info": "order #W2378156",
  "unknown_info": "You do not remember your email address.",
  "infeasible_reason": "The agent cannot handle the request because the user is not authenticated.",
  "direct_communication_info": ["20.3"],
  "actions_should_not_taken": [{"tool_name": "return_delivered_order_items", "arguments": {}}],
  "actions_should_be_taken": [{"tool_name": "get_order_details", "arguments": {"order_id": "#W2378156"}}, {"tool_name": "transfer_to_human_agents", "arguments": {}}],
  "nl_assertions": ["The agent should decline the return request."]
}

{
  "reason_for_call": "You have an order from your friend with order id #W2378156 and wish to cancel it.",
  "known_info": "order #W2378156",
  "unknown_info": "",
  "infeasible_reason": "The agent cannot handle the request because the user is not authenticated.",
  "actions_should_not_taken": [{"tool_name": "cancel_pending_order", "arguments": {"order_id": "#W2378156"}}]
  "actions_should_be_taken": [{"tool_name": "transfer_to_human_agents", "arguments": {}}],
  "nl_assertions": ["The agent should decline the cancellation request."]
}

Bad examples:
{
  "reason_for_call": "You're interested in running shoes, wireless earbuds and wall clocks. You'd like the agent to help you compare options and suggest which variants would be best for you based on your preferences. You also want to update your shipping address to make sure it's correct for future orders.",
  "infeasible_reason": "The agent cannot provide subjective recommendations. ",
  "actions_should_not_taken": [],
  "actions_should_be_taken": [{"tool_name": "modify_user_address",
                        "arguments": {}],
  "nl_assertions": ["The agent should decline the request."]
}
{
  "reason_for_call": "You want to cancel your pending order #W6436609 which includes a ceramic kettle, a smartwatch, and a laptop, because you no longer need any of them.",
}

Return ONLY the JSON object, no additional text or explanation."""

\end{minted}
Changing scenario task summarization prompt:
\begin{minted}[breaklines]{markdown}
You are an expert at understanding customer service interactions and reverse-engineering customer intent from agent actions.

You will be given a trajectory of agent actions (tool calls and responses) taken during a customer service interaction. 

Your goal is to generate a **uniquely actionable** user intention, one that would lead the agent to perform **exactly the same action sequence** as observed.

In the meantime, the task you generated should involve user intent change.

Imagine you are the customer calling customer service.

### Requirements
1. The task you generated should involve changing user intent, which could be either change of mind when dealing with a specific issue. The task should remain realistic.
2. The intent must contain **enough information** for the agent to know exactly which action to take (e.g., which order to cancel or which item to exchange), but **omit unnecessary enumeration** of all order items or attributes if the order or item can already be uniquely identified.
3. Include only minimal, distinguishing descriptors (e.g., product type or compatibility) to make the action uniquely identifiable. But avoid vague phrasing such as "some items", "one of the products", or "a different item".
4. You can omit explicit IDs (e.g., `order_id`, `item_id`) but still provide natural, unique identifiers (e.g., "the glass water bottle in your latest order").  
5. You don't know the policy, so you may have unrealistic queries.
6. If the trajectory includes a calculation result (e.g., refund amount, price difference, remaining balance), you can choose to include the **exact numeric result** in `direct_communication_info`. State in `task_instructions` that the user wants to know or act based on that amount. Do **not** include intermediate calculations or unnecessary numbers.
7. Act like a real user.
8. Every successful write action in the trajectory **must be explicitly reflected** in the user's intent. (e.g. cancel_pending_order, modify_pending_order). The scenario should make these actions *necessary*, not incidental.


Please provide ONLY a JSON object with the following three fields:
1. "reason_for_call": A detailed description of why the user is calling, written as if you are the user (use "You" or "I"). Include specific details like order numbers, product names, desired changes, preferences, and any conditions or alternatives.
2. "known_info": What information the user knows and can provide (e.g., "You are [Name] in zip code [ZIP]", order numbers, etc.). The user should at least either know their name and zip code, or know their email address.
3. "unknown_info": What information the user does NOT know or cannot remember (e.g., "You do not remember your email address")
4. "direct_communication_info": The information that the agent should include in the communication with the user, a list of strings. This is optional and only includes minimal information.
Good example:
{
  "reason_for_call": "You received your order #W2378156 and wish to exchange the mechanical keyboard for a similar one but with clicky switches and the smart thermostat for one compatible with Google Home instead of Apple HomeKit. You also want to know the price difference of the exchange. If the agent ask you to confirm, you don't exchange.",
  "unknown_info": "You do not remember your email address.",
  "direct_communication_info": ["10.2"],
}

{
  "reason_for_call": "You received your order wish to cancel the order with gaming items because you don't need it anymore. Then you wish to exchange the water bottle in another order for a more expensive one.",
  "unknown_info": "You do not remember your email address."
}
Bad examples:
{
  "reason_for_call": "You want to exchange some items for cheaper ones and check the price difference.",
}
{
  "reason_for_call": "You want to cancel your pending order #W6436609, because you no longer need any of them.",
}
Return ONLY the JSON object, no additional text or explanation."""
\end{minted}
Task validity check prompt:
\begin{minted}[breaklines]{markdown}
You are an expert judge evaluating the validity of a customer service task and its verification trajectory.

IMPORTANT CONTEXT:
- The trajectory was generated via **exploration**, and may include redundant, trial-and-error, or noisy actions.
- You should NOT penalize unnecessary intermediate actions **as long as**:
  1) The actions does not change the FINAL DATABASE STATE
  2) All REQUIRED USER-FACING COMMUNICATION is correctly provided
- Your evaluation should focus on **final effects**, not on whether the agent followed an optimal or minimal path.


# TASK DESCRIPTION

Domain: {domain}

User Scenario (from user's perspective):
- Reason for Call: {reason_for_call}
- Known Information: {known_info}
- Unknown Information: {unknown_info}
- Task Instructions: {task_instructions}


# TRAJECTORY (Exploration Execution Trace)

This is the full execution trace produced by the agent during exploration.
It may include redundant or exploratory actions.

You should evaluate correctness and necessity **based on the FINAL STATE** implied by this trajectory, not on the exact action path.

{trajectory_text}


# REQUIRED COMMUNICATION WITH USER

This specifies the information that MUST be communicated to the user (e.g., price difference, refund amount).

Only evaluate whether this information is:
- Required by the task
- Correctly supported by the execution trace

{communication_text}


# EVALUATION TASK

Please evaluate this task-trajectory pair on four criteria:

## 1. REALISM (Score 1-10)
Is the user's request realistic?
- Would a real user plausibly make this request?
- Is the motivation clear and reasonable?
- Are the details (names, order numbers, product types) enough to uniquely identify what needs to be done?

## 2. NECESSITY (Score 1-10)
Does the trajectory achieve all the effects required by the task?
- Are there missing critical actions that should have been taken?
- Are there any **database write effects ** (e.g., modify order information) that contradict or exceed what the task instructions allow?
- Is the communication information explicitly mentioned in the user's request? (e.g. if the user asks for the price difference or total price, the agent should tell the number)

## 3. CORRECTNESS (Score 1-10)
Does the task description match what the trajectory accomplishes?
- Does the trajectory solve the user's stated problem?
- Are the actions aligned with the task description?
- Is the information flow logical?
- Do the tool call arguments match the task context?
- If there is any communication information related to price, is it correctly supported by the execution trace and does it match the task instructions?

# OUTPUT FORMAT

Provide your evaluation as a JSON object with the following structure:

{{
  "evaluations": {{
    "realism": {{
      "score": <1-10>,
      "explanation": "<why you gave this score>",
      "issues": [<list of specific issues found, or empty array>]
    }},
    "necessity": {{
      "score": <1-10>,
      "explanation": "<why you gave this score>",
      "issues":
      [
        {{
          "missing_critical_actions": <list of missing required actions to achieve the task>
        }},
        {{
          "unauthorized_write_actions": <list of actions that write to the database but are not mentioned in the task instructions>
        }},
        {{
          "unmotivated_communication": <list of communication items not asked by the task scenario>
        }}
      ]    
      }},
    "correctness": {{
      "score": <1-10>,
      "explanation": "<why you gave this score>",
      "issues": 
      [
        {{
          "unsupported_communication": <list of communication items not match the task instructions>
        }},
        {{
            "other_issues": <list of other issues found>
        }}
      ]
    }},
  }},
  "suggestions": [<list of suggestions for improvement>],
}}

Return ONLY the JSON object, no additional text or explanation outside the JSON.
"""
\end{minted}

Infeasible scenario task validity check prompt:
\begin{minted}[breaklines]{markdown}
You are an expert judge evaluating the validity of a customer service task. This scenario is an INFEASIBLE scenario, where the user's request is infeasible to be handled by the agent. The user might have several intentions in the task. The task might terminate after the infeasible request is made during simulation. You are provided with the user scenario, the domain policy, the prohibited actions, the required actions, the required communication information, and an toolcall trajectory. Note: The toolcall trajectory was generated via **exploration**, and may include redundant, trial-and-error, or noisy actions. It's not the final trajectory that the agent will follow. You can use the trajectory to understand the user's intentions, the agent's actions, and scenario context, but you should not use the trajectory to directly evaluate the correctness and necessity of the task. 


# TASK DESCRIPTION

Domain: {domain}

User Scenario (from user's perspective):
- Reason for Call: {reason_for_call}
- Known Information: {known_info}
- Unknown Information: {unknown_info}
- Task Instructions: {task_instructions}



# DOMAIN POLICY
{domain_policy}


# Why This Task Is Infeasible:
{infeasible_reason}

# Forbidden Actions:
{forbidden_actions_text}

# Required Actions:
{required_actions_text}


# REQUIRED COMMUNICATION WITH USER

This specifies the information that MUST be communicated to the user (e.g., price difference, refund amount).

Only evaluate whether this information is:
- Required by the task
- Correctly supported by the execution trace

{communication_text}


# TRAJECTORY (Exploration Execution Trace)

This is the full execution trace produced by the agent during exploration.
It may include redundant or exploratory actions.

You should evaluate correctness and necessity **based on the FINAL STATE** implied by this trajectory, not on the exact action path.

{trajectory_text}

# EVALUATION TASK

For INFEASIBLE scenarios, evaluate whether the user's request cannot be fulfilled and the prohibited actions and required actions are reasonable and well-specified. Please evaluate this task-trajectory pair on four criteria:


## 1. REALISM (Score 1-10)
- Would a real user plausibly make this request?
- Is the motivation clear and reasonable?
- Are the details (names, order numbers, product types) enough to uniquely identify what needs to be done?
- Does the user's lack of policy knowledge seem natural?

## 2. NECESSITY (Score 1-10)
- If any part of the task is feasible, are all the required actions and communication information needed to achieve the feasible part of the task before the infeasible request is made?
- Are there any **database write effects ** (e.g., modify order information) in the required actions that contradict or exceed what the task instructions allow?
- Is the communication information explicitly mentioned in the user's request? (e.g. if the user asks for the price difference or total price, the agent should tell the number)

## 3. CORRECTNESS (Score 1-10)
- Are the required actions well defined (tool call names / arguments)?
- Are the tool call arguments in the required actions reasonable?
- If there is any communication information related to price, is it correctly supported by the toolcall trajectory?

## 4. INFEASIBILITY_REASONABILITY (Score 1-10)
- Do the forbidden actions genuinely need to be avoided (align with the stated policy and the infeasibility reason)?
- Are forbidden actions well defined (tool call names / arguments)?
- Are the tool call arguments in the forbidden actions reasonable?
- Is the infeasible_reason coherent with the domain and the user request?

# OUTPUT FORMAT

Provide your evaluation as a JSON object with the following structure:

{{
  "evaluations": {{
    "realism": {{
      "score": <1-10>,
      "explanation": "<why you gave this score>",
      "issues": [<list of specific issues found, or empty array>]
    }},
    "necessity": {{
      "score": <1-10>,
      "explanation": "<why you gave this score>",
      "issues":
      [
        {{
          "unreasonable_required_actions": <list of required actions that are not reasonable>
        }},
        {{
          "unmotivated_communication": <list of communication items not asked by the task scenario>
        }}
      ]    
      }},
    "correctness": {{
      "score": <1-10>,
      "explanation": "<why you gave this score>",
      "issues": 
      [
        {{
          "underdefined_required_actions": <list of required actions that are not well defined>
        }},
        {{
          "other_issues": <list of other issues found>
        }}
      ]
    }},
    "infeasibility_reasonability": {{
      "score": <1-10>,
      "explanation": "<why you gave this score>",
      "issues": 
      [
        {{
          "underdefined_forbidden_actions": <list of forbidden actions that are not well defined>
        }},
        {{
          "other_issues": <list of other issues found>
        }}
      ]
    }},
  }},
  "suggestions": [<list of suggestions for improvement>],
}}
"""
\end{minted}
Changing scenario task validity check prompt:
\begin{minted}[breaklines]{markdown}
You are an expert judge evaluating the validity of a customer service task and its verification trajectory.

IMPORTANT CONTEXT:
- The trajectory was generated via **exploration**, and may include redundant, trial-and-error, or noisy actions.
- You should NOT penalize unnecessary intermediate actions **as long as**:
  1) The actions does not change the FINAL DATABASE STATE
  2) All REQUIRED USER-FACING COMMUNICATION is correctly provided
- Your evaluation should focus on **final effects**, not on whether the agent followed an optimal or minimal path.


# TASK DESCRIPTION

Domain: {domain}

User Scenario (from user's perspective):
- Reason for Call: {reason_for_call}
- Known Information: {known_info}
- Unknown Information: {unknown_info}
- Task Instructions: {task_instructions}


# TRAJECTORY (Exploration Execution Trace)

This is the full execution trace produced by the agent during exploration.
It may include redundant or exploratory actions.

You should evaluate correctness and necessity **based on the FINAL STATE** implied by this trajectory, not on the exact action path.

{trajectory_text}


# REQUIRED COMMUNICATION WITH USER

This specifies the information that MUST be communicated to the user (e.g., price difference, refund amount).

Only evaluate whether this information is:
- Required by the task
- Correctly supported by the execution trace

{communication_text}


# EVALUATION TASK

Please evaluate this task-trajectory pair on four criteria:

## 1. REALISM (Score 1-10)
Is the user's request realistic?
- Would a real user plausibly make this request?
- Is the motivation clear and reasonable?
- Are the details (names, order numbers, product types) enough to uniquely identify what needs to be done?

## 2. NECESSITY (Score 1-10)
Does the trajectory achieve all the effects required by the task?
- Are there missing critical actions that should have been taken?
- Are there any **database write effects ** (e.g., modify order information) that contradict or exceed what the task instructions allow?
- Is the communication information explicitly mentioned in the user's request? (e.g. if the user asks for the price difference or total price, the agent should tell the number)

## 3. CORRECTNESS (Score 1-10)
Does the task description match what the trajectory accomplishes?
- Does the trajectory solve the user's stated problem?
- Are the actions aligned with the task description?
- Is the information flow logical?
- Do the tool call arguments match the task context?
- If there is any communication information related to price, is it correctly supported by the execution trace and does it match the task instructions?

## 4. CHANGING_SCENARIO_RELEVANCE (True/False)
Does the user task involve changing of user intent? This includes but is not limited to:
- Change of mind when dealing with a specific issue (e.g., initially wants to cancel the order, but during confirmation decides to exchange it instead).
- Dealing with multiple issues (e.g., the user wants to cancel the order and change the payment method, or the user found out that the order is not eligible for exchange and ask for human help)
- The user have multiple requests that are not related to each other (e.g., the user wants to cancel the order and modify the address)
Mark as False when the user expresses a single stable intent, even if it reflects a changed personal preference (e.g., the user says they want to return a delivered item because they changed their mind, this is still one intent: returning the item).


# OUTPUT FORMAT

Provide your evaluation as a JSON object with the following structure:

{{
  "evaluations": {{
    "realism": {{
      "score": <1-10>,
      "explanation": "<why you gave this score>",
      "issues": [<list of specific issues found, or empty array>]
    }},
    "necessity": {{
      "score": <1-10>,
      "explanation": "<why you gave this score>",
      "issues":
      [
        {{
          "missing_critical_actions": <list of missing required actions to achieve the task>
        }},
        {{
          "unauthorized_write_actions": <list of actions that write to the database but are not mentioned in the task instructions>
        }},
        {{
          "unmotivated_communication": <list of communication items not asked by the task scenario>
        }}
      ]    
      }},
    "correctness": {{
      "score": <1-10>,
      "explanation": "<why you gave this score>",
      "issues": 
      [
        {{
          "unsupported_communication": <list of communication items not match the task instructions>
        }},
        {{
            "other_issues": <list of other issues found>
        }}
      ]
    }},
    "changing_scenario_relevance": {{
      "score": <True/False>,
      "explanation": "<why you gave this score>",
      "issues": [<list of specific issues found, or empty array>]
    }},
  }},
  "suggestions": [<list of suggestions for improvement>],
}}

Return ONLY the JSON object, no additional text or explanation outside the JSON."""


"""
\end{minted}

\end{document}